\documentclass[conference]{IEEEtran}
\IEEEoverridecommandlockouts
\usepackage{cite}
\usepackage{amsmath,amssymb,amsfonts}
\usepackage{algorithmic}
\usepackage{graphicx}
\usepackage{textcomp}
\usepackage{xcolor}
\def\BibTeX{{\rm B\kern-.05em{\sc i\kern-.025em b}\kern-.08em
    T\kern-.1667em\lower.7ex\hbox{E}\kern-.125emX}}

\usepackage[colorlinks=true]{hyperref}
\usepackage[dvipsnames]{xcolor}
\hypersetup{
    colorlinks=true,
    linkcolor=black,
    citecolor=black,
    urlcolor=blue!70!black
}

\usepackage{listings}
\usepackage{multicol}
\usepackage{parskip}
\usepackage{titlesec}
\usepackage{graphicx}
\usepackage{enumitem}
\usepackage{lipsum}  % Remove when no longer needed

\usepackage{xspace}

\newcommand{\appname}{\texttt{OnPrem.LLM}\xspace}

\begin{document}

\title{Generative AI for FFRDCs}

\author{\IEEEauthorblockN{Arun S. Maiya}
\IEEEauthorblockA{\textit{~}
\textit{Institute for Defense Analyses (IDA)}\\
Alexandria, VA, USA \\
amaiya@ida.org}
}

\maketitle

\begin{abstract}
Federally funded research and development centers (FFRDCs) face text-heavy workloads, from policy documents to scientific and engineering papers, that are slow to analyze manually. We show how large language models can accelerate summarization, classification, extraction, and sense-making with only a few input-output examples. To enable use in sensitive government contexts, we apply OnPrem.LLM, an open-source framework for secure and flexible application of generative AI. Case studies on defense policy documents and scientific corpora, including the National Defense Authorization Act (NDAA) and National Science Foundation (NSF) Awards, demonstrate how this approach enhances oversight and strategic analysis while maintaining auditability and data sovereignty.
\end{abstract}

\begin{IEEEkeywords}
generative AI, large language models, LLMs
\end{IEEEkeywords}

\section{Introduction}
Federally Funded Research and Development Centers (FFRDCs) are institutions that conduct research and development (R\&D) and provide independent, policy-relevant analyses to support government decision-making, oversight, and strategic planning. Generative AI and large language models (LLMs) offer FFRDCs unprecedented capabilities for analyzing policy, summarizing documents, and surfacing insights.
                                                                                                                                                This paper concisely outlines concrete use cases, critical risks, and strategies for deploying LLMs responsibly and practically in oversight and institutional contexts. Our intended audience is researchers from diverse fields with only modest coding skills and little or no machine learning background.

\section{Motivation}
Generative AI is particularly well-suited for FFRDC work:

\begin{itemize}
    \item Policy analysis and strategic planning often involve complex, text-heavy workflows and limited labeled data due to sensitivity and compartmentalization. LLMs can tackle such tasks using only a few input-output demonstrations (i.e., few-shot prompting) \cite{b1}.
    \item Generative AI tools have evolved to work more easily with locally stored data in air-gapped networks \cite{b2}.
    \item Local, open-weight models enable secure on-premises deployment for sensitive document data \cite{touvron2023llamaopenefficientfoundation}.
\end{itemize}                                                                                                                                                  

\section{OnPrem.LLM}
All applications of generative AI in this paper are implemented using \appname, a privacy-conscious Python toolkit for document intelligence \cite{b2}. \appname is open source, free to use under the Apache 2.0 license, and available
on GitHub at \url{https://github.com/amaiya/onprem}.
                                                                                                                                                
\appname is organized into four primary modules that together provide a comprehensive framework for document intelligence, which we now describe.

\subsection{LLM Module}
The {\em LLM} module is the core engine for interfacing with large language models. It provides a unified API for working with various LLM backends including llama.cpp, Hugging Face Transformers, Ollama, vLLM, and a wide-range of cloud providers  \cite{anthropic2024claude3,llama_cpp_2023,ollama2025,kwon2023efficient,openai2024gpt4technicalreport,wolf2020huggingfacestransformersstateoftheartnatural}. This module abstracts the complexity of different model implementations through a consistent interface while handling critical operations such as model loading with inflight quantization support, easy accessibility to LLMs served through APIs, agentic-like retrieval-augmented generation (RAG), and structured LLM outputs.

\subsection{Ingest Module}
The {\em Ingest} module is a comprehensive document processing pipeline that transforms raw documents into retrievable knowledge. It supports multiple document formats with specialized loaders, automated OCR for image-based text, and extraction of tables from PDFs. The module offers three distinct vector storage approaches:

\begin{enumerate}
\item \textbf{Dense Store}: Implements semantic search using sentence transformer embeddings and ChromaDB for similarity-based retrieval using hierarchical navigable small-world (HNSW) indexes \cite{chroma2025}. Elasticsearch is also supported.
\item \textbf{Sparse Store}: Provides both on-the-fly semantic search and traditional keyword search through Whoosh\footnote{We use \texttt{whoosh-reloaded} (available at \url{https://github.com/Sygil-Dev/whoosh-reloaded}), a fork and continuation of the original Whoosh project. } (or Elasticsearch) with custom analyzers and custom fields.
\item \textbf{Dual Store}: Combines both approaches by maintaining parallel stores, enabling hybrid retrieval that leverages both semantic similarity and keyword (or field) matching.
\end{enumerate}

\subsection{Pipelines Module}
The {\em Pipelines} module includes pre-built workflows for common document intelligence tasks with specialized submodules:

\begin{itemize}
\item \textbf{Extractor}: Applies prompts to document units (sentences/paragraphs/passages) in order to extract structured information with Pydantic model validation \cite{pydantic2025}.
\item \textbf{Summarizer}: Provides document summarization with multiple strategies including map-reduce for large documents and concept-focused summarization.
\item \textbf{Classifier}: Implements text classification through wrappers for scikit-learn (\texttt{SKClassifier}), Hugging Face transformers (\texttt{HFClassifier}), and few-shot learning with limited examples (\texttt{FewShotClassifier})   \cite{pedregosa2011scikit,tunstall2022efficientfewshotlearningprompts,wolf2020huggingfacestransformersstateoftheartnatural}.
\item \textbf{Agent}: Builds LLM-powered agents to execute complex tasks using tools and custom, user-defined methods.
\end{itemize}

\subsection{App Module}
\begin{figure}[htbp]
\centerline{\includegraphics[width=\columnwidth]{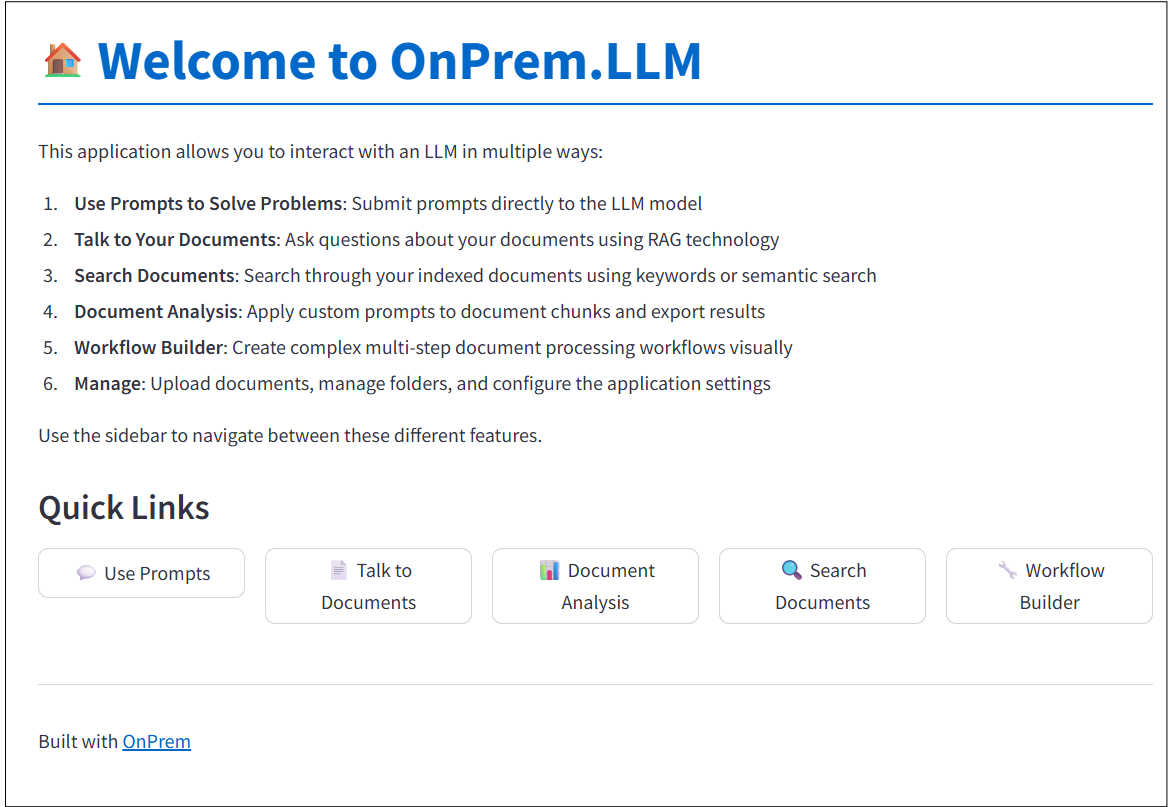}}
\caption{The \appname Web UI.}
  \label{fig:webui}

\end{figure}

As shown in Figure \ref{fig:webui}, a Streamlit-based web application makes the system accessible to non-technical users through six specialized interfaces \cite{streamlit2025}. The web interfaces offer easy, point-and-click access to: 1) interactive chat with conversation history; 2) document-based question answering with source attribution to mitigate hallucinations; 3) keyword and semantic search with filtering, pagination, and result highlighting; 4) custom prompt application to individual document passages with Excel export capabilities; 5) visual workflow builder for complex document analysis pipelines; and 6) an administrative interface for document ingestion, folder management, and application configuration. More information in Appendix B.

\section{Example Use Cases}                                                                                                                 
In this section, we explore four concrete applications of generative AI. The examples highlight use cases common to FFRDCs, particularly study and analysis centers. All examples rely on publicly available data that are easily accessible to readers, and primarily focus on demonstrating the capabilities and versatility of the {\em LLM} module in \appname.

\subsection{Summarization: Technology-Focused Synopses}
General summaries of documents are not always useful. Rather, government analysts are often more interested in summarizing documents with       respect to a specific concept of interest. We refer to this as {\em concept-focused summarization}. For instance, lengthy government documents  can be summarized with respect to specific technology areas of interest.  
~\\
{\em Code Example Using NDAA:} \url{https://amaiya.github.io/onprem/examples_summarization.html}.
                                                                                                                                                \subsection{Classification: Open-Ended Surveys}
FFRDCs are often tasked with analyzing free-text responses to open-ended survey instruments, including public comments. For instance,           governments use Requests for Information (RFIs) to solicit input on a wide range of scientific, technical, and policy issues. Large language    models (LLMs) can reduce the time needed to analyze and {\em code} RFI responses---a process of classifying responses into categories---from weeks to minutes.
~\\
{\em Code Example Using RFI Responses}: \url{https://amaiya.github.io/onprem/examples_qualitative_survey_analysis.html}
                                                                                                                                                                                                                                                                                                \subsection{Extraction: Legal and Regulatory Analysis}
LLMs excel at information extraction with very few ground truth examples \cite{b1}. For instance, LLMs can identify which portions of the Federal Acquisition Regulation (FAR) are driven by statutory requirement by identifying and extracting references to statutes. 
~\\
{\em Code Example Using Federal Acquisition Regulation}: \url{https://amaiya.github.io/onprem/examples_legal_analysis.html}

\subsection{Question-Answering: Horizon Scanning}
Semantic search retrieves information based on meaning rather than exact keyword matches and is commonly used in LLM-powered question-answering systems such as retrieval-augmented generation (RAG) \cite{b3}. Such techniques are valuable for horizon scanning—the systematic       analysis of research grants, publications, and other sources to identify emerging trends, opportunities, and potential disruptions in science   and technology. This process can help surface early signals of significant developments across research landscapes.
~\\
{\em Code Example Using NSF Awards Data:} \url{https://amaiya.github.io/onprem/examples_rag.html}

\section{Risks and Misuses}
\begin{itemize}
   \item \textbf{Hallucination}: LLMs can produce confident, fluent output that is factually incorrect or misleading \cite{b4}.
\item \textbf{Over-reliance}: Analysts must remain critical of AI-generated content and avoid uncritical acceptance of opaque model reasoning.  \item \textbf{Data Exposure}: Use of cloud-based LLMs requires strict safeguards to prevent inadvertent disclosure of sensitive or proprietary  information \cite{b5}.
\end{itemize}

\section{Mitigation Strategies}
\begin{itemize}                                                                                                                                     \item \textbf{Auditability}: Implement workflows and guardrails that enhance transparency and traceability, enabling analysts to verify     outputs and detect hallucinations or extraction errors (e.g., automated checks comparing outputs to inputs) \cite{b6}. See use of \texttt{attempt\_fix} in Appendix A.
                                                                                                                                                    \item \textbf{Human-in-the-Loop}: Depending on task risk, ensure domain experts remain actively involved in AI-assisted workflows to        prevent over-reliance and maintain expert oversight (e.g., reviewing cited and scored sources of AI-generated answers).                                                                                                                                                                             \item \textbf{Sovereign Deployment}: Prefer on-premises LLM deployments when handling sensitive, proprietary, or mission-critical data to   reduce exposure risk, retain control, and process data where it already resides. Leverage cloud as permitted or needed. See Appendix A for a quick start example.

\end{itemize}

\section{Conclusion}                                                                                                                        While not a silver bullet, generative AI—when implemented with clear institutional values and rigorous oversight—can be a powerful tool for     advancing FFRDC missions. More information and examples at \url{https://amaiya.github.io/onprem/}.

\bibliographystyle{IEEEtran}
\bibliography{IEEEabrv,main}

\section*{Appendix A: Quick Start Code Example}
\label{appendix:quickstart}
\appname supports a wide range of LLM backends, including llama.cpp, Hugging Face Transformers, Ollama, vLLM, and cloud providers such as     OpenAI, Anthropic, and Amazon Bedrock.
                                                                                                                                                In this example, Ollama (local) and Anthropic (cloud) are used as the LLM backends. Ollama can be downloaded from \url{https://ollama.com/},    and an API key for Anthropic’s cloud service is available at \url{https://console.anthropic.com/}. 
 
{\footnotesize                                                                                                                                               
\begin{lstlisting}
# install
!pip install onprem[chroma]
from onprem import LLM, utils

# local LLM with Ollama as backend
!ollama pull llama3.1
llm = LLM('ollama/llama3.1')

# basic prompting
result = llm.prompt('Give me a short one sentence definition of an LLM.')

# RAG
utils.download(
   'https://www.arxiv.org/pdf/2505.07672',
   '/tmp/my_documents/paper.pdf')
llm.ingest('/tmp/my_documents')
result = llm.ask('What is OnPrem.LLM?')

# switch to Anthropic Claude as backend
llm =LLM('anthropic/claude-3-7-sonnet-latest')

# information extraction w/ structured outputs
from pydantic import BaseModel, Field
class MeasuredQuantity(BaseModel):
  value: str = Field(description="numerical value")
  unit: str = Field(description="unit of measurement")
result = llm.pydantic_prompt(
           'He was going 35 mph.',
           attempt_fix=True,
           pydantic_model=MeasuredQuantity)
print(result.value) # 35
print(result.unit)  # mph

\end{lstlisting}
}

Please refer to the \appname documentation (\url{https://amaiya.github.io/onprem/}) for more information and examples.                                                                                                                                                    

\section*{Appendix B: The Web UI}
\label{appendix:webui}

\appname comes pre-packaged with a web-based user interface (UI).  The web UI currently consists of six main screens that allow users to apply LLMs using a point-and-click interface with no coding required.

\begin{figure}[htbp]
\centerline{\includegraphics[width=\columnwidth]{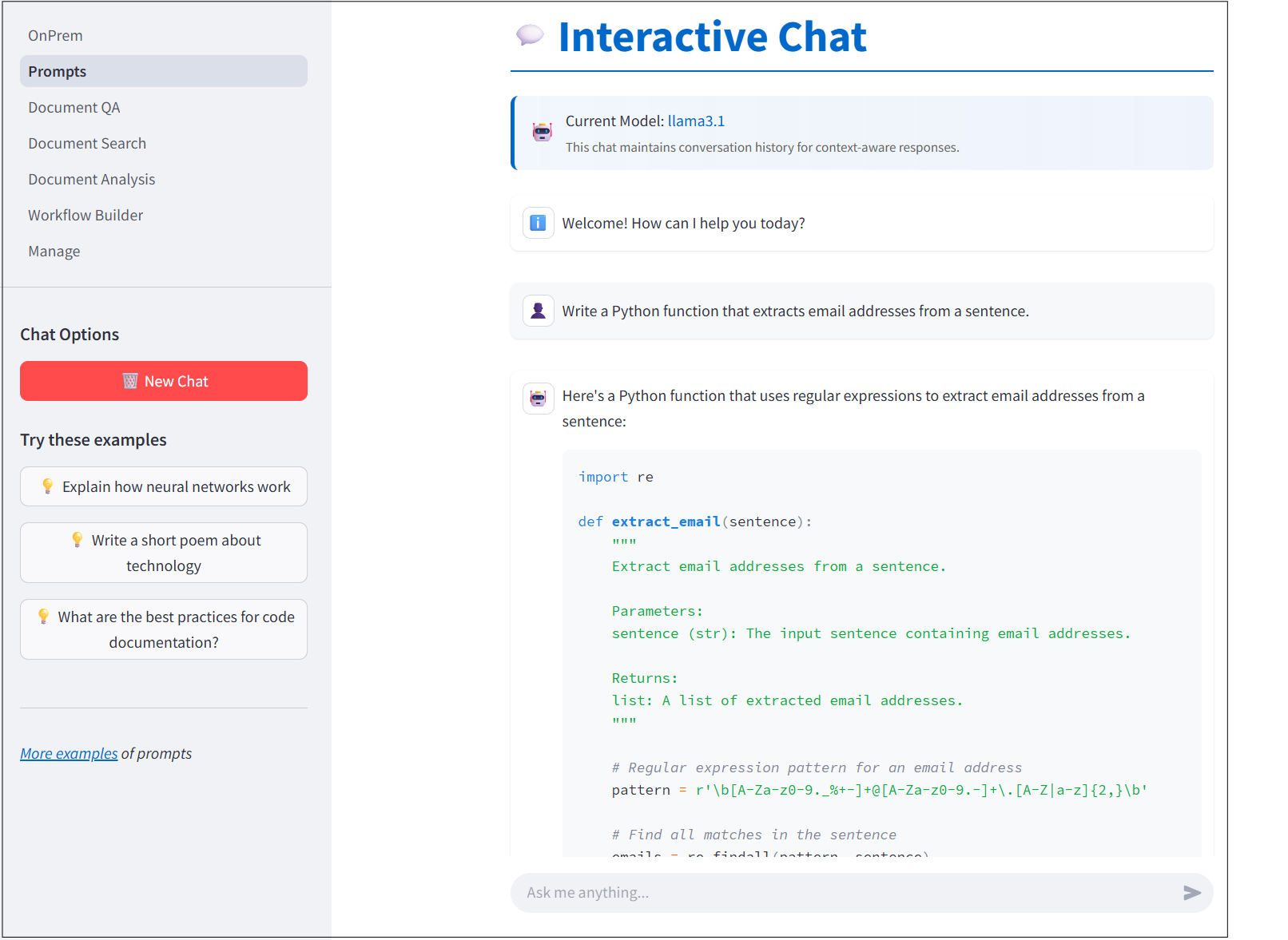}}
\caption{Interactive Chat.}
  \label{fig:prompts}

\end{figure}

{\bf 1. Prompts.} The {\em Prompts} screen, shown in Figure \ref{fig:prompts}, allows users to interactively chat with an LLM while maintaining conversational history similar to other consumer-facing chat bots like ChatGPT.

\begin{figure}[htbp]
\centerline{\includegraphics[width=\columnwidth]{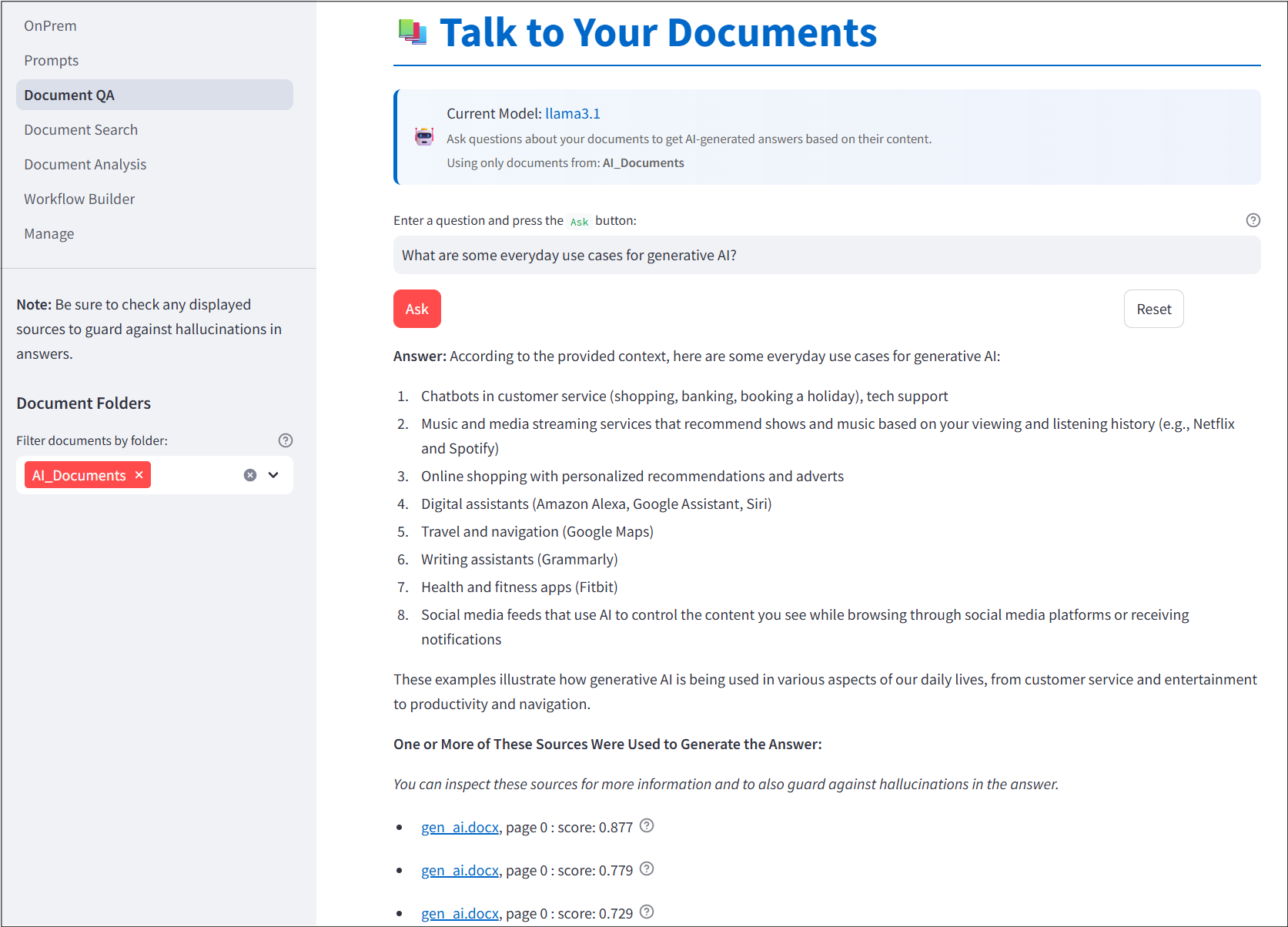}}
\caption{Document Question-Answering.}
  \label{fig:qa}

\end{figure}

{\bf 2. Document QA.} The {\em Document QA} screen, shown in Figure \ref{fig:qa}, allows users to submit questions to an LLM and generate cited answers from uploaded documents using RAG. 

\begin{figure}[htbp]
\centerline{\includegraphics[width=\columnwidth]{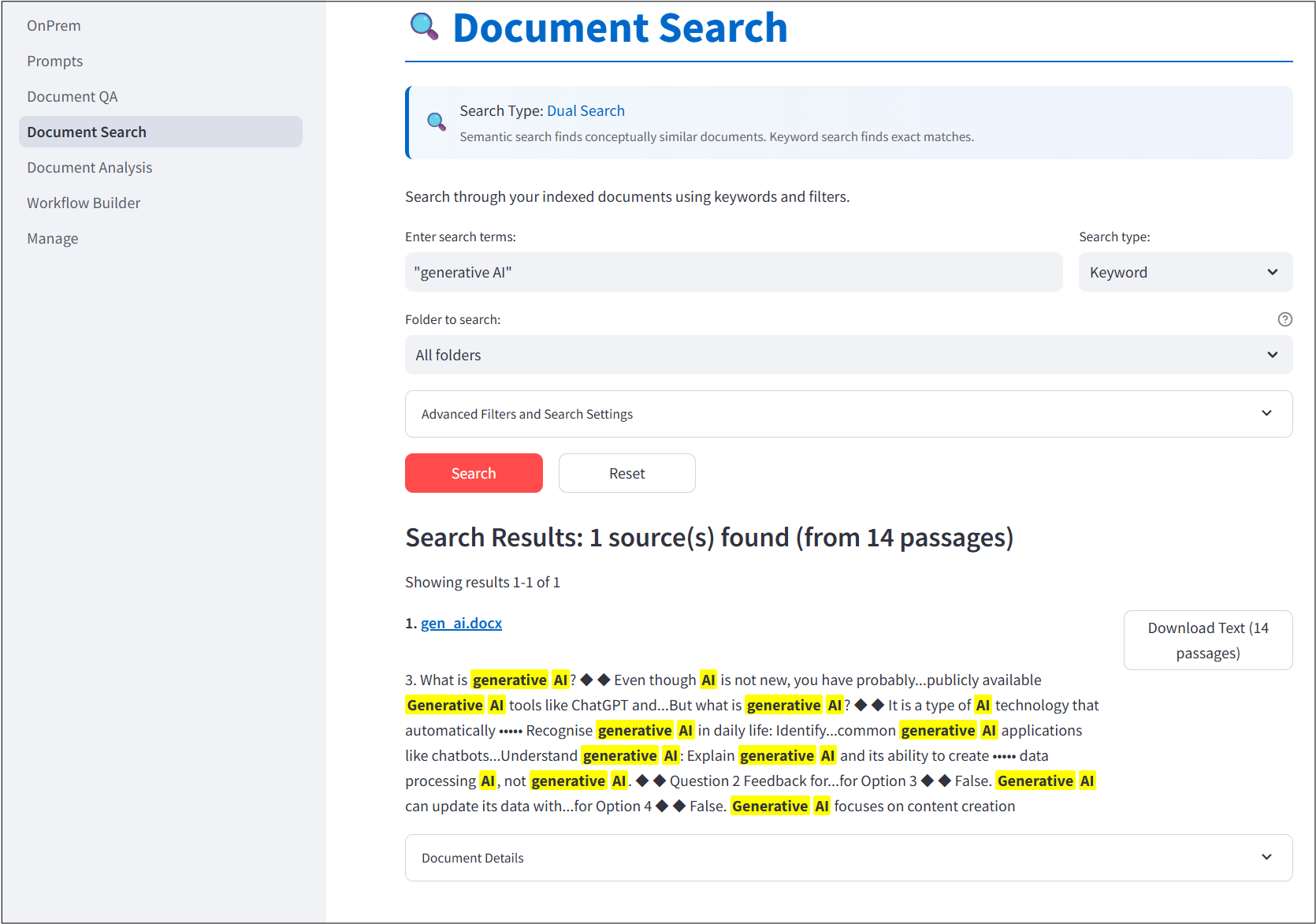}}
\caption{Document Search.}
  \label{fig:search}

\end{figure}

{\bf 3. Document Search.} In addition to answering questions about documents, it is often useful to identify documents or passages within documents that match key  phrases or a boolean query.  The {\em Document Search} screen in Figure \ref{fig:search} supports exactly this.

{\bf 4. Document Analysis.} The {\em Document Analysis} screen allows users to apply a prompt to every passage in a document and is useful for information extraction tasks. Screenshot omitted due to space constraints.

\begin{figure}[htbp]
\centerline{\includegraphics[width=\columnwidth]{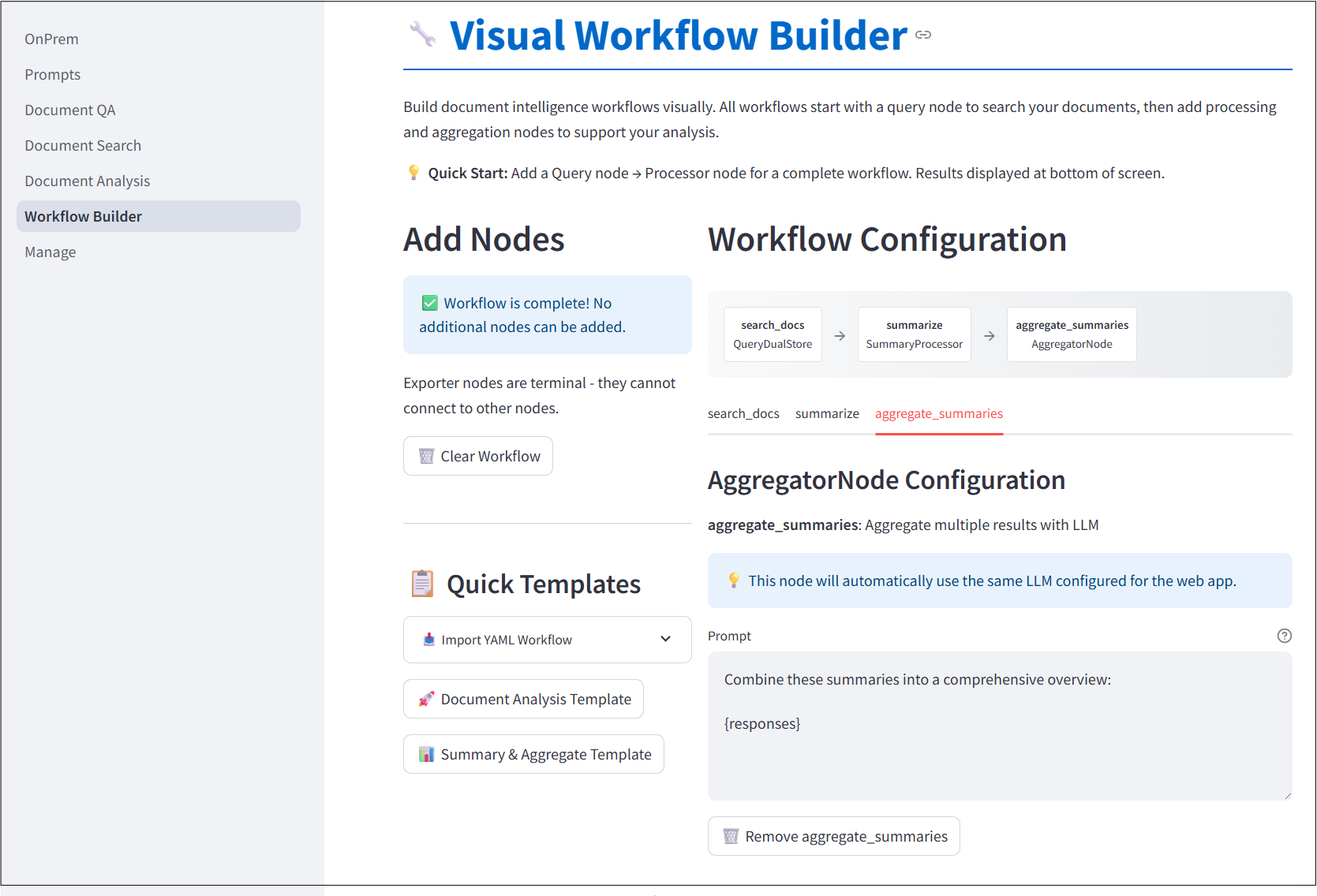}}
\caption{Visual Workflow Builder.}
  \label{fig:workflow}
\end{figure}

{\bf 5. Visual Workflow Builder.} The {\em Workflow Builder}, shown in Figure \ref{fig:workflow}, provides a point-and-click interface for creating complex document analysis pipelines.
  
{\bf 6. Document Management and Settings.} Many of the aforementioned screens assume the availability of a repository of ingested documents. The {\em Manage} interface is used to manage settings and document repositories and upload documents for analyses. Screenshot is omitted due to space constraints.

For more information, please see the web UI documentation: \url{https://amaiya.github.io/onprem/webapp.html}.

%\begin{figure}[htbp]
%\centerline{\includegraphics[width=\columnwidth]{figures/onprem_upload.png}}
%\caption{Document Management.}
%  \label{fig:upload}
%\end{figure}

\section*{Appendix C:\\Ethical Use of Data and Informed Consent}
\label{appendix:ethics}
This work did not involve human participants or subjects. All data used were publicly available.
%~\\
%{\em Ethical Use of Data and Informed Consent.}
%~\\
%This work did not involve human participants or subjects. All data used were publicly available.

\end{document}